\documentclass[11pt]{article}

\usepackage[preprint]{acl}

\usepackage{times}
\usepackage{latexsym}

\usepackage[T1]{fontenc}

\usepackage[utf8]{inputenc}

\usepackage{microtype}

\usepackage{inconsolata}

\usepackage{graphicx}

\usepackage{hyperref}
\usepackage{url}
\usepackage{booktabs}
\usepackage{svg}
\usepackage{tabularx}
\usepackage{fancyvrb}
\usepackage{fvextra}
\usepackage{subcaption}

\usepackage{float}

\title{Masked diffusion LLMs can use EoS tokens for hidden reasoning}

\author{Sarah Breckner\textsuperscript{1,2}  \hspace{2cm} Sebastian Schuster\textsuperscript{1}\\ \ \\
\textsuperscript{1}Faculty of Computer Science, University of Vienna, Vienna, Austria \\
\textsuperscript{2}UniVie Doctoral School Computer Science,
University of Vienna,
Vienna, Austria \\ \ \\
\texttt{\{sarah.breckner,sebastian.schuster\}@univie.ac.at}}

\begin{document}
\maketitle

\begin{abstract}
Diffusion LLMs have been proposed as an alternative to autoregressive LLMs. Curiously, they are especially capable if the generation length, i.e., the number of tokens the model has to output, is set to a much higher value than the correct answer length, and the model pads its answer with end-of-sequence (EoS) tokens. We hypothesize that off-the-shelf masked diffusion LLMs use the representations of EoS tokens as additional computing capacity, which enhances their performance. We experiment with the diffusion models LLaDA1.5, LLaDA2.0-mini, and Dream-v0 on three reasoning tasks: Addition, Entity Tracking, and Sudoku. In a controlled prompting experiment, we confirm that adding EoS tokens improves the LLMs' performance. To further verify whether their representations are used for hidden computations, we perform a causal intervention and transfer the hidden states of the EoS tokens between generations, which increases the models' relative likelihood of outputting the counterfactual answer. The behavioral experiments and the causal interventions indicate that fully bidirectional masked diffusion LLMs can indeed perform latent reasoning in the representations of EoS tokens. Furthermore, we find that these results generalize beyond toy tasks and that providing the model with additional EoS tokens also improves performance on GSM8K and two-hop reasoning.
\end{abstract}

\maketitle
 
\section{Introduction}
Inspired by their success in computer vision, diffusion models have emerged as an alternative to autoregressive large language models (LLMs). They perform particularly well at complex reasoning tasks with interdependent sub-goals, such as Sudoku \citep{ye2025beyond}. However, the underlying reasons for their improved performance remain largely an open question.

Diffusion language models (LMs) and autoregressive LMs differ along several dimensions. First, akin to masked LMs \citep{devlin-etal-2019-bert}, diffusion models use bidirectional attention, which allows them to construct representations of the entire input. Second, while autoregressive LMs decode one token at a time, diffusion models jointly decode a set of masked tokens. After each generation step, some tokens are retained, and the rest are re-masked until all masks have received their assignment.

\begin{figure}[t]
    \centering
    \includegraphics[width=0.8\linewidth]{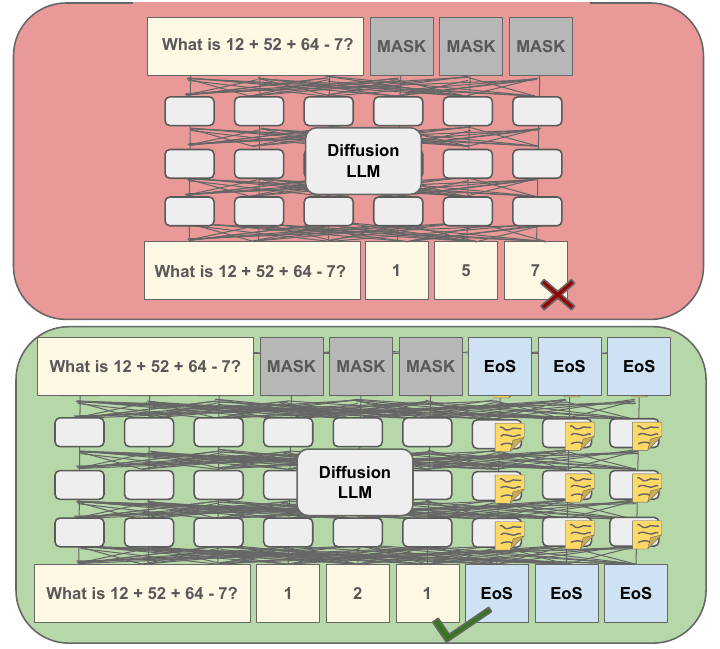}
    \caption{LLMs use the hidden representations of the prompt and the output tokens for computations (top). When increasing the generation length of diffusion LLMs, they pad excess positions with EoS tokens and thereby increase their reasoning budget, which leads to better task performance (bottom).}
    \label{fig:one}
\end{figure}

This flexible generation order, however, does not allow for variable output lengths and requires users to pre-set the number of tokens to generate. To nonetheless allow for shorter generations, the models predict end-of-sequence (EoS) tokens for the excess positions.  Interestingly, we observe that a higher generation length can yield a higher accuracy with the same number of generated contentful tokens padded with more EoS tokens. Moreover, previous work indicates that the models place high attention weights on EoS tokens \citep{chen2026dpad}.

These observations lead us to hypothesize that off-the-shelf masked diffusion LLMs use the hidden representations of EoS tokens as an additional space for reasoning during inference, as illustrated in Figure~\ref{fig:one}. Unlike overt reasoning, such as chain-of-thought (CoT), where the model generates intermediate steps that add to both the context and the computational budget, latent reasoning utilizes only the extra compute and occurs entirely within the hidden representations \citep{pfau2024let} as a form of test-time scaling 
\citep{muennighoff-etal-2025-s1}.

In this paper, we test this hypothesis and investigate through behavioral and intervention experiments whether the capability for latent reasoning in the hidden representations of EoS tokens has emerged in off-the-shelf diffusion LLMs. We find that this is the case, though not surprisingly, this phenomenon is primarily limited to fully bidirectional models and largely absent from models with block-causal attention.

Our contributions are as follows:
\begin{enumerate}
    \item We establish a relationship between generation length and model performance through behavioral experiments showing that increasing the generation length of diffusion LLMs improves performance (Section~\ref{exp1}).
    \item We disentangle the effect of the number of decoding steps and the number of EoS tokens through a controlled experiment in which we equip the model with different numbers of fixed EoS tokens (Section~\ref{exp2}).
    \item We demonstrate that the representations of the EoS tokens causally affect the generation of the response through a causal intervention experiment (Section~\ref{exp3}).
    \item We show that these findings also extend to naturalistic tasks such as mathematical problem solving and two-hop reasoning (Section~\ref{exp4}). 
\end{enumerate}

\section{Background and Related Work}
\paragraph{Diffusion models} were initially developed for image generation \citep{ho2020denoising, song2021denoising}. The diffusion process establishes a training trajectory by stepwise adding noise to an image. Then, the model is trained to reverse this path and step-by-step decode noise into an image. 

Adapting diffusion models for language modeling tasks, i.e., for the prediction of discrete tokens, requires a discrete diffusion process \citep{austin2021structured, hoogeboom2021argmax, sahoo2024simpleeffectivemaskeddiffusion}. One way to achieve this is through masking
\citep{bie2025llada20scalingdiffusionlanguage, nie2025large, ye2025dream7bdiffusionlarge}. Masked diffusion LMs encode the text with a bidirectional transformer. Instead of adding noise to semantic representations, variable amounts of the training data are masked, and the model is trained to reconstruct the missing tokens.
\paragraph{Latent reasoning} is a phenomenon in language models where their reasoning is not happening overtly, unlike in the case of generated textual reasoning traces, such as a CoT or a scratchpad \citep{wei2022chain, nye2021workscratchpadsintermediatecomputation}. One way to realize latent reasoning is through token repetitions that provide additional compute but no semantic information, e.g., periods or whitespaces \citep{bavandpour2025lower}. However, under which conditions LLMs exhibit such reasoning is an active area of research. Smaller autoregressive models have to be trained from scratch or finetuned to introduce the reasoning tokens \citep{kshitij_sachan_llms_2023, goyal2024think, herel2024thinking, pfau2024let, wang2024guiding}. In contrast, very recent frontier models are able to perform latent reasoning on a range of filler tokens that are inserted into the assistant's answer before generation \citep{ryan_greenblatt_recent_2025, ryan_greenblatt_recent_2026, baherwani2026llmreasoningvisiblechainofthought} and their representations of filler tokens contain decodable intermediates of multi-hop problems \citep{brauer2026reading}. For masked diffusion LMs, it has so far only been shown that latent reasoning is possible through training from scratch with a specific reasoning objective \citep{he2026reasoninglatenttokensdiffusion}. Whether and under which conditions latent reasoning also emerges from the standard pre- and post-training objectives of diffusion LMs is the focus of our work.

\section{Experimental Setup}
\begin{figure*}[t]
    \centering
    \includegraphics[width=0.9\linewidth]{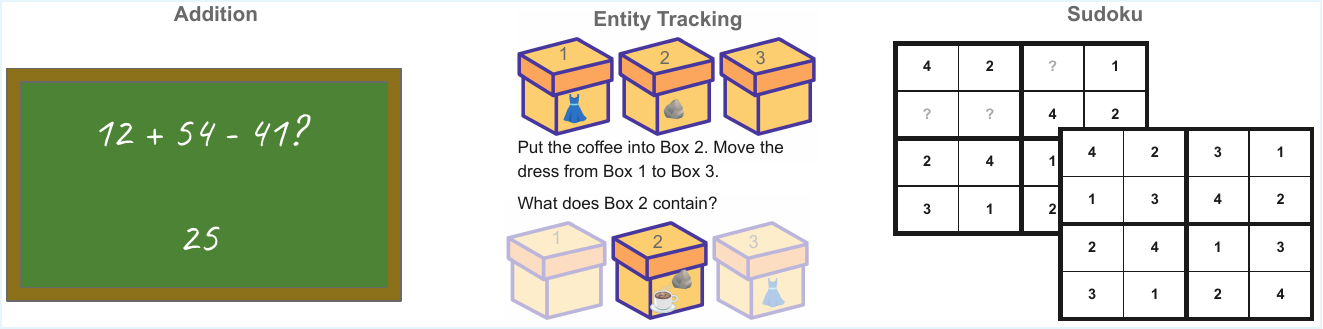}
    \caption{Illustration of the tasks.}
    \label{fig:datasets}
\end{figure*}

\paragraph{Datasets} \label{section:datasets}
In our experiments, we examine LLM reasoning without CoT, and therefore, we select representative tasks that are simple enough to be solved without a verbose reasoning trace. We use datasets of the domains math, natural language understanding, and planning; examples are shown in Figure~\ref{fig:datasets}. This provides a systematic framework with scalable problem difficulty and constrained output lengths. To complement this setup, we apply our controlled prompting experiment to two less systematic but more naturalistic settings, GSM8K \citep{cobbe2021gsm8k} and two-hop reasoning \citep{biran-etal-2024-hopping}. 

As math data, we generate addition and subtraction problems of integers from 0 to 100 with a positive result. We increase the difficulty from two to six operands, with 200 examples each. To test the models' capacity to follow a natural language text, we utilize the entity tracking dataset introduced by \citet{kim-schuster-2023-entity}. It contains descriptions of the contents of boxes and operations that move, remove, and add objects. The model is asked to predict the final state of one box. A data point is characterized by the total number of operations in the description, as well as how many of them concern the target box. The dataset is balanced across the combination of these two measures. Lastly, we choose 4x4 Sudoku as a representative of an emblematic planning problem for diffusion models \citep{ye2025beyond}. We vary the difficulty by increasing the number of empty cells from one to twelve and generate 200 examples each \citep{Sudoku4LLM}. 

In addition, we verify our hypothesis on the test set of GSM8K, which consists of textual math problems with solutions that require reasoning over multiple steps \citep{cobbe2021gsm8k}, and on a two-hop question answering dataset \citep{biran-etal-2024-hopping}. It comprises descriptions such as \textit{the spouse of the composer of The Blue Danube is}. To find the described entity, the model first has to resolve the composer, Johann Strauss, to then name his spouse, Angelika Strauss. For each model, we use a custom test set of the two-hop problems, whose one-hop parts the model correctly solves in isolation. Thereby, we ensure that the focus remains on its reasoning capabilities rather than fact retention. The datasets and code are available on Github,\footnote{\url{https://github.com/minaolensarah/thinking_eos-by-eos}}, and detailed dataset statistics and prompts are provided in Appendix~\ref{sec:appendix:prompts}. 
\paragraph{Models} 
We investigate three instruction-tuned diffusion LLMs: the two smaller models, LLaDA1.5 \citep{nie2025large} and Dream-v0 \citep{ye2025dream7bdiffusionlarge} with 8B and 7B parameters, respectively, and the mixture-of-experts LLaDA2.0-mini \citep{bie2025llada20scalingdiffusionlanguage} with 16B parameters. 

The post-training phase for LLaDA1.5 and Dream-v0 is very similar; they are both instruction-tuned on sequences padded with EoS tokens that are not masked, so that the model learns to generate adequately short answers. Both models use fully bidirectional attention, allowing early positions to attend to the padded EoS tokens, which could enable them to learn using EoS tokens for hidden reasoning.

In contrast, some recent diffusion models, such as LLaDA2.0, are fine-tuned using block-wise generation \citep{arriola2025block}. There, the generation window is segmented, and tokens are generated in arbitrary order within a block, but blocks are filled sequentially. Furthermore, LLaDA2.0 is optimized during finetuning for block-causal attention: all tokens within one block can attend to each other and to earlier blocks. In our experiments, we set the block size equal to the generation length, which disables block-wise generation and allows the models to attend to the trailing EoS tokens and potentially use them for hidden reasoning.

These differences in fine-tuning and decoding procedures between diffusion models allow us to investigate under which conditions reasoning in latent representations emerges in diffusion language models. Finally, to compare the results to autoregressive LLMs, we also test the 8B parameter, instruction-tuned versions of Llama3.1 \citep{grattafiori2024llama} and Qwen3 \citep{yang2025qwen3technicalreport} without reasoning tokens as baselines. 

\section{Experiment~1 -- Prompting }  \label{exp1}
\begin{figure*}[t]
    \centering
    \includegraphics[width=0.85\linewidth]{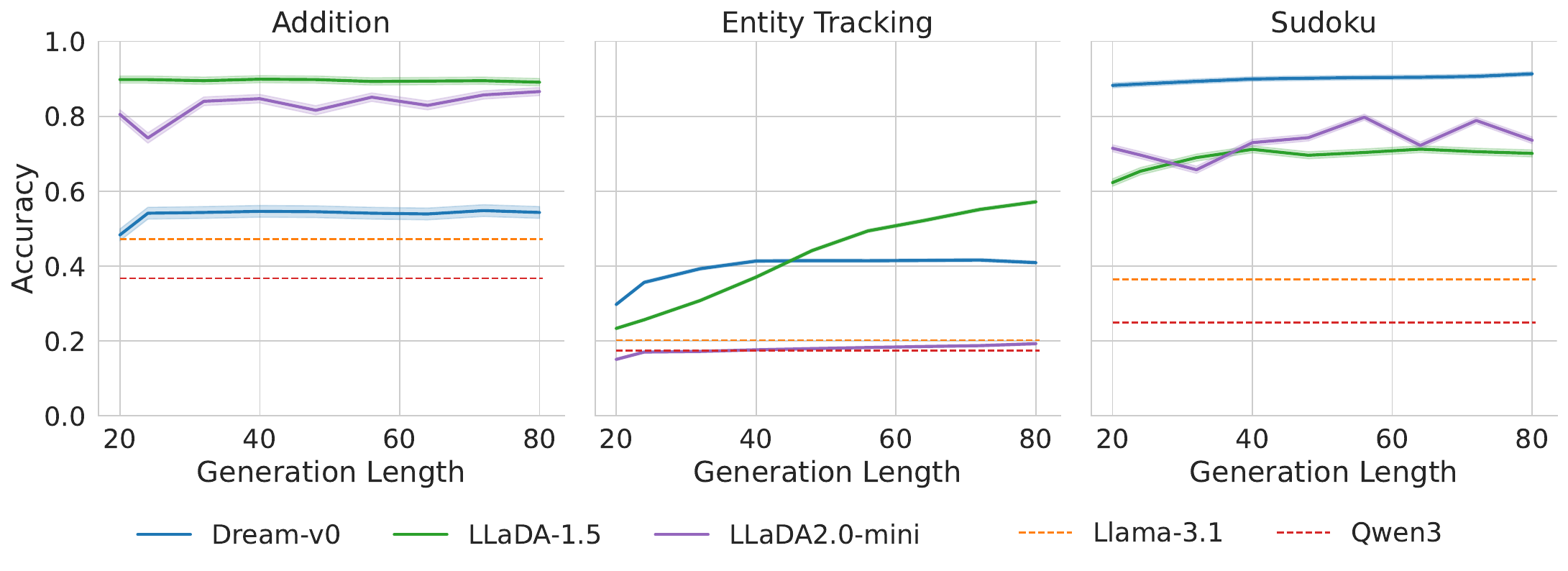} 
    \caption{Aggregate results of the prompting experiment (Exp.~1). The accuracy of the diffusion models with standard errors for generation lengths higher than required to fit the answer. For comparison, we show the performance of the autoregressive LLMs as dashed lines.}
    \label{fig:observations:all_accs}
\end{figure*}
\begin{figure*}[t]
    \centering
    \includegraphics[width=0.75\linewidth]{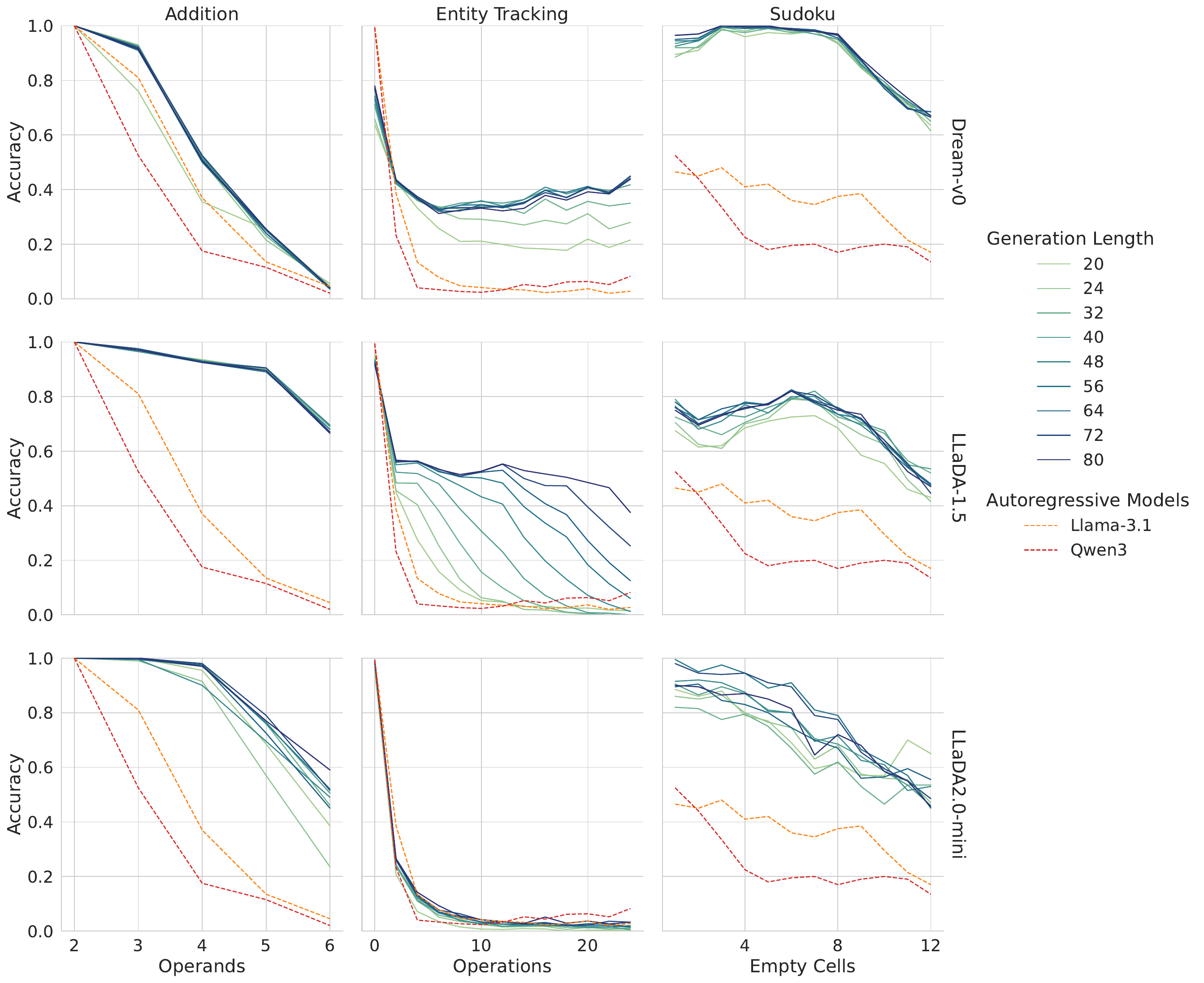}    
    \caption{Fine-grained results of the prompting experiment (Exp.~1). The accuracy of the diffusion models across task-specific difficulty levels and generation lengths. The autoregressive LLMs are marked as dashed lines.}
    \label{fig:observations:acc_over_diff}
\end{figure*}
We first investigate how the generation length affects the performance across three diffusion LLMs (LLaDA1.5, LLaDA2.0-mini, and Dream-v0) and compare them to two autoregressive baselines (Llama3.1 and Qwen3). While we increase the total number of tokens to generate, we fix the number of tokens to retain per step. 

\subsection{Results}
The aggregate accuracies are shown in Figure~\ref{fig:observations:all_accs}. For Dream and LLaDA1.5, the performance mostly increases with a higher generation length until a point of saturation, after which it plateaus or slightly declines. In contrast, LLaDA2.0 displays unstable performance with an upward trend. Across models, LLaDA1.5 on \textsc{Entity Tracking} has the greatest improvement from 23\% to 67\% from generation length 20 to 80.

In addition, we analyze the models' performance in relation to the problem difficulty, which is displayed in Figure~\ref{fig:observations:acc_over_diff}. Here, we generally observe one of two trends. For many models and tasks, the model performance degrades less with task difficulty when the generation length is increased. For example, most strikingly, both LLaDA1.5 and Dream show considerably improved performance for more challenging examples of \textit{Entity Tracking} when a high generation length is used. For other model-task combinations (e.g., Dream or LLaDA2.0-mini on \textsc{Addition}), there is a small increase in performance on more difficult problems when going from a generation length of 20 to slightly higher generation lengths. However, beyond that, performance appears to be saturated and no longer increases with generation length, as also observed in the aggregate results (Fig.~\ref{fig:observations:all_accs}).

The results indicate a dependency between the generation length and the task performance, even though the shortest generation window already fits the correct answer. Analyzing the generations in more detail, we further find that the number of contentful tokens remains constant or declines when the generation length increases (see Figure~\ref{fig:observations:length} in the Appendix). Inversely, the number of padded EoS tokens grows to fill the remaining positions. 
\begin{figure*}[t]
    \centering
    \includegraphics[width=0.8\linewidth]{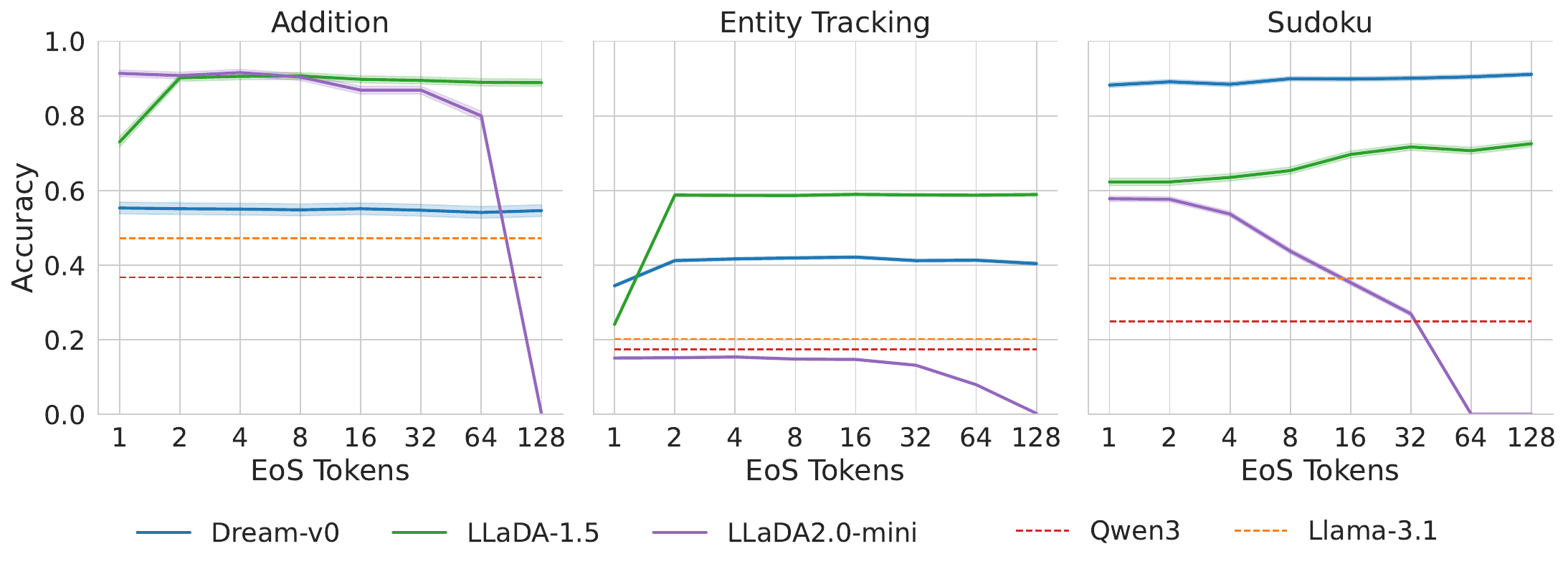}
    \caption{Aggregate results of the controlled prompting experiment (Exp.~2) where we pad an increasing number of EoS tokens in the start state. The accuracy is shown with standard error, and the autoregressive LLMs are marked with dashed lines.}
    \label{fig:eos:all_accs}
\end{figure*}
\subsection{Discussion}
In this experiment, we investigate the effect of the generation length on the masked diffusion model performance. Overall, an increase in generation length results in an increase in accuracy up until a saturation point, and it leads to a more stable performance on more difficult data points.

Here, the generation length is tied to both the number of decoding steps and the number of EoS tokens, which increase when the model has more positions to fill. Both factors could positively impact the performance. On the one hand, it has been established that more decoding steps lead to better predictions \citep{nie2025large, ye2025dream7bdiffusionlarge}. On the other hand, we hypothesize that diffusion models use EoS tokens for latent reasoning. The more EoS tokens are available, the larger the number of representations that could be used as additional compute. This might enable the LLM to solve more complex tasks.  To disentangle the effect of the decoding steps and the EoS tokens, we design a setup in which we can keep the decoding steps constant while varying the number of trailing EoS tokens.

\section{Experiment~2 -- Controlled Prompting}  \label{exp2}
We find that the accuracy of the masked diffusion models increases with the generation length, and identify two possible causes for the performance improvement, namely the number of decoding steps and the number of padded EoS tokens, which both grow with the generation length. To disentangle their influence, we perform a controlled prompting experiment, keeping decoding steps constant and manually increasing the number of EoS tokens. We achieve this by starting the stepwise generation from a fixed number of masks, followed by a variable number of EoS tokens. With this setup, we can examine if EoS tokens influence the task performance of the diffusion models.  
\subsection{Results}
We append an increasing number of EoS tokens and fix the number of masks and decoding steps. Figure~\ref{fig:eos:all_accs} shows the accuracies in relation to the number of EoS tokens. On the \textsc{Entity Tracking} dataset, Dream and LLaDA1.5 exhibit a jump in performance from 1 to 2 additional EoS tokens. Beyond that, adding more EoS tokens increases the accuracy only by a small amount. Conversely, on \textsc{Sudoku}, the two models improve steadily. Until 128 EoS tokens, Dream improves by 3\% and LLaDA1.5 by 10\%. This is contrasted by LLaDA2.0, which almost does not benefit from trailing EoS tokens. When appending 64 or 128 EoS tokens, its generation even breaks, and it outputs only EoS tokens.
\subsection{Discussion}
In this experiment, we systematically increase the number of padded EoS tokens with constant decoding steps and study the model performance. This serves as an indicator of whether diffusion LLMs use the EoS tokens to support their reasoning process. For the fully bidirectional models, we find a positive effect of appending EoS tokens. In some cases, the improvement is steady; in others, it mirrors a jump followed by a plateau. This differs from the previous experiment, and might be due to implicit length regularization by the padded start state. For example, on \textsc{Entity Tracking}, LLaDA1.5 reaches an accuracy of 57\% at generation length 80 in the previous experiment, while it outperforms this score by 2\% already with two padded EoS tokens, i.e., a generation length of 24. This might be because the lower number of masks prevents the common error of predicting too many objects.

LLaDA2.0's generation breaks down when a large number of EoS tokens are appended. These observations contrast with the previous experiment, in which its performance improves with the generation length, and it does not produce broken outputs. There, the model first fixes some EoS tokens and the scaffolding of the solution, and then stepwise generates the answer. In contrast, the start state in this experiment consists of masks followed by EoS tokens as if the model had already fixed tokens right-to-left. While this deviation from its semi-autoregressive generation method was intentional to allow the model to attend to the EoS tokens at every decoding step, it negatively affects the performance, and the results indicate that LLaDA2.0 is unable to leverage these trailing tokens as additional computational budget. 

LLaDA2.0’s sensitivity to trailing EoS chains highlights a fundamental tension: while additional tokens can provide additional compute, they also alter the output distribution, which influences which subsequent tokens the model can generate. While LLaDA1.5 and Dream seem robust against long EoS chains, these sequences degrade LLaDA2.0's performance. A similar trade-off becomes clear in our experiments with alternative reasoning fillers (see Appendix \ref{otherTokens_exp2}). While any token could theoretically serve for latent reasoning, its practical utility depends on whether it remains in-distribution. For LLaDA1.5 and Dream, EoS tokens offer the benefit of extra compute without distorting the output distribution, which does not hold for LLaDA2.0.

For the two fully bidirectional models, the results support the hypothesis that EoS tokens are used for latent reasoning, but how quickly their benefits saturate depends on the task.  On two out of the three datasets, the simplest way to achieve close to the best performance is to append 2 EoS tokens to the start state.
\section{Experiment~3 -- Causal Intervention} \label{exp3}
\begin{figure*}[t]
    \centering
    \includegraphics[width=0.78\linewidth]{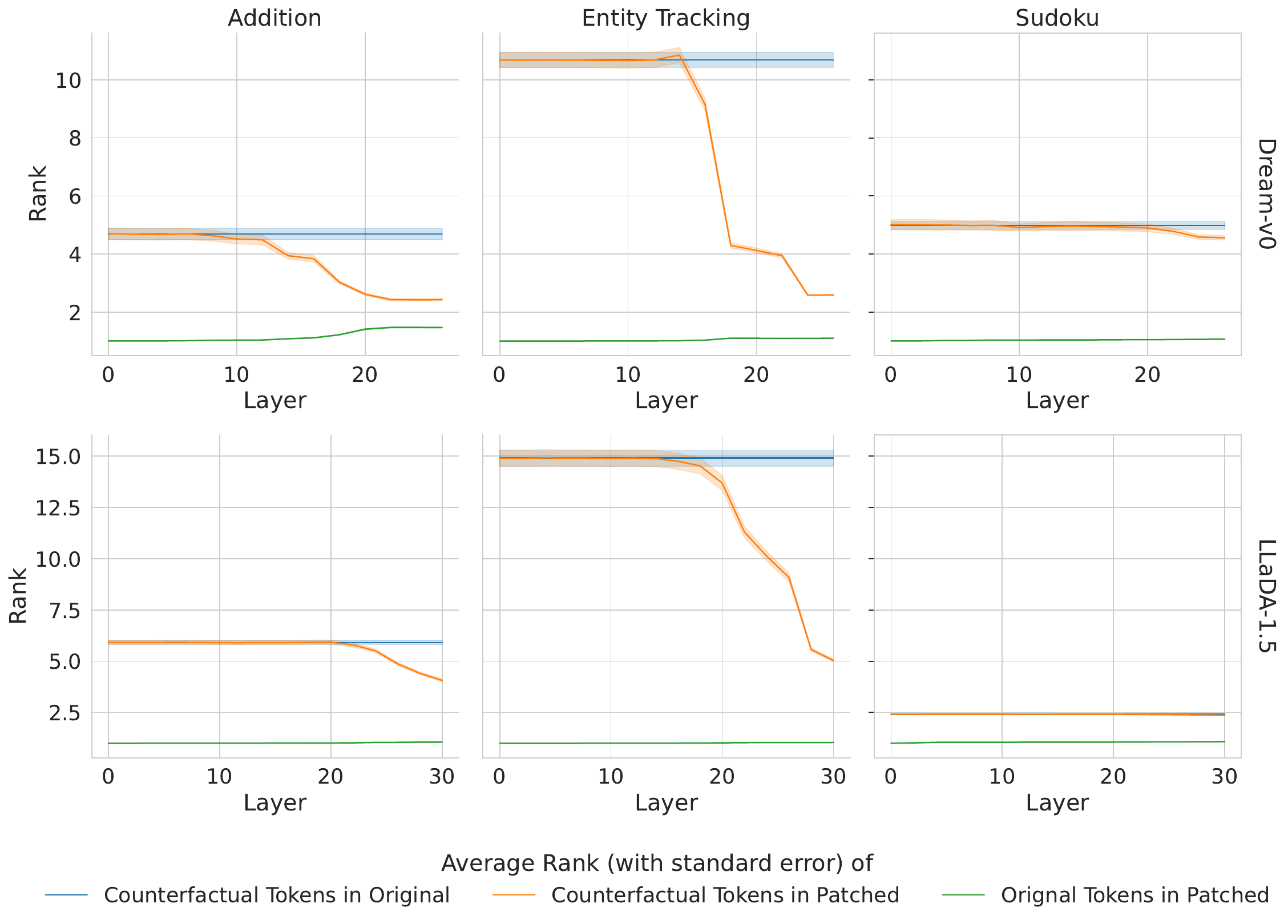}
    \caption{Results of the causal intervention (Exp.~3). The difference between the rank of the counterfactual tokens in the original generation (blue) and their rank after patching (orange) indicates how much the patched representations affect the output distribution. A decrease suggests that the hidden states contain latent computations.}
    \label{fig:patching}
\end{figure*}
So far, we have examined behaviorally how the reasoning performance of diffusion LLMs is tied to the generation length and, more concretely, to the number of trailing EoS tokens. The results suggest that fully bidirectional diffusion LLMs use the EoS tokens for hidden computations. To provide further support for this hypothesis, we perform a causal intervention experiment in which we transfer the hidden states of the EoS tokens in one generation with those of the EoS tokens of a counterfactual generation \citep{geiger2021causalabstractions, zhang2024towards}. For example, in the \textsc{Entity Tracking} task, we pair problems with the same description but query for different boxes. Then, we transfer all hidden states of the EoS tokens until a target layer to evaluate the impact on the output distribution. We measure the counterfactual and original tokens' average rank, where a decrease in rank of the counterfactual tokens means an increase in relative likelihood, which signifies that the patching transferred latent computations. If, on the contrary, the EoS tokens are nothing but padding, exchanging them should have little to no systematic impact. 

As a counterfactual in the \textsc{Addition} task, we swap the operators: plus to minus and vice versa. For the \textsc{Entity Tracking} scenarios, we use the same description but ask for the contents of a different box. For \textsc{Sudoku}, we keep the empty cells and perform a circular shift to the other numbers. For this experiment, we do not apply stepwise decoding, since otherwise, the EoS tokens could contain lexical information on answer tokens from earlier decoding steps. Further, we do not perform the interventions on LLaDA2.0 as it is a mixture-of-experts architecture in which a different subset of experts is activated for each token, making activation patching infeasible. As a control, we repeat the activation patching with other potential reasoning tokens. These results are shown in Appendix \ref{otherTokens_exp3}.

\subsection{Results}
Figure \ref{fig:patching} shows the average rank of the output tokens before and after the causal intervention. We exclude cases in which the counterfactual generation matches the original one, for example, if the model falsely predicts that both boxes are empty, and cases in which it only generates filler tokens for the masked positions in the counterfactual run, because such instances are not insightful on whether the hidden states contain task-relevant information. 

For both models, the causal intervention leads to a considerable decrease in the rank of the counterfactual tokens for the \textsc{Addition} and \textsc{Entity Tracking} tasks. Dream on \textsc{Addition} has the earliest impact of the patching already at layer 10, whereas for most task-model combinations, we see effects only starting in the middle layers. The patching on \textsc{Sudoku} has a smaller impact. The overall low ranks of the tokens for LLaDA1.5 indicate that it mostly predicts empty cells for both the original and counterfactual runs. This might be due to the one-step decoding. Dream seems more robust to the highly parallel decoding, and the intervention leads to a small decrease in rank. 

Across the three datasets and two models, we find that the intervention considerably changes the output distribution, which suggests that the models can use the hidden states of the padded EoS tokens as extra reasoning budget, though this might not always become evident in a single decoding step.

\section{Experiment~4 -- Naturalistic Tasks}  \label{exp4}
\begin{figure*}[t]
\includegraphics[width=0.85\linewidth]{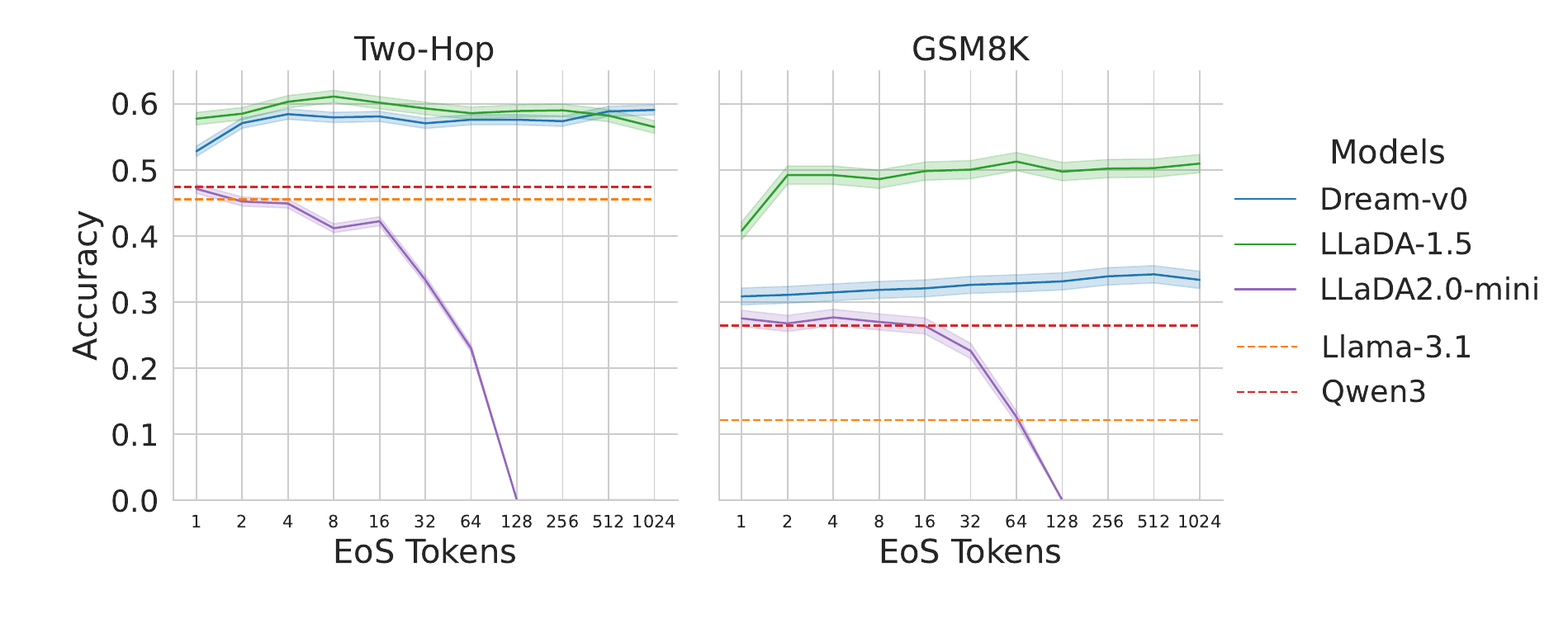}
        \caption{Accuracy with standard error on more naturalistic datasets, two-hop and GSM8K (Exp.~4). As for Experiment~2, we pad the start state with an increasing number of EoS tokens. For comparison, we show the performance of the autoregressive LLMs as dashed lines.}
    \label{fig:gsm8k}
\end{figure*}
In the previous experiments, we collected evidence that masked diffusion LLMs use EoS tokens for hidden reasoning. To do so, we used controlled datasets that allow for balancing the difficulty of the data and controlling the output lengths. This, however, comes at the cost of naturalness. Therefore, we validate our findings on two more naturalistic datasets of math (GSM8K, \citep{cobbe2021gsm8k}) and multi-hop reasoning (two-hop, \citep{biran-etal-2024-hopping}) using the setup of Experiment~2. We fix the number of masks and decoding steps to suffice for the expected output and add an increasing number of EoS tokens at the end of the start state to monitor whether this impacts the model performance. This indicates whether the models can use EoS tokens to extend their reasoning budget in more naturalistic settings, too.

\subsection{Results}
On the more naturalistic datasets, the three models display similar patterns as on the controlled ones in Experiment~2. The results are shown in Figure~\ref{fig:gsm8k}. Dream on \textsc{two-hop} and LLaDA1.5 on \textsc{GSM8K} show a jump in performance from one to two added EoS tokens. In the two remaining cases, the two models show a steadier improvement in performance but a small decline after a peak is reached. The number of EoS tokens that yield the best performance depends on the task and model. As in the previous experiment, LLaDA2.0 is not robust towards trailing EoS tokens and its performance drops to zero, providing additional evidence that the latent reasoning behavior is limited to fully bidirectional diffusion language models.

\section{Conclusion}
Diffusion models perform better than autoregressive LLMs on several reasoning tasks, but there remain open questions on why exactly this is the case. In this work, we investigated the improved model performance with increased generation length, which mostly results in more EoS tokens. We then tested through behavioral and intervention experiments whether masked diffusion LLMs can use EoS representations for additional hidden computations. Behaviorally, we found that for the majority of tasks, the presence of EoS tokens leads to better performance, suggesting that they are used for latent reasoning. This is present in diffusion models with fully bidirectional attention, whereas the model optimized for block-causal attention shows unstable improvements with generation length and seems negatively affected by long EoS chains in its initial state. In addition, we performed a causal intervention on the hidden states of the EoS tokens. This decreased the rank of the counterfactual tokens, underscoring that EoS tokens carry task-relevant information. Lastly, we confirmed that the benefits of added EoS tokens for reasoning persist on two more naturalistic datasets.

For models without block-causal attention, there seem to be two methods for unlocking the capacity for hidden reasoning: either by choosing a higher generation length than required or by padding the start state with EoS tokens. However, increasing the generation length comes at the cost of more computations and decoding steps. If, instead, one appends EoS tokens, we found that for some tasks and models under examination, appending 2 EoS tokens yields close to the best results while requiring fewer decoding steps and maintaining a short output length.  

These results mirror the latent reasoning behavior of autoregressive LMs, which, on smaller models, has to be introduced through careful training \citep{goyal2024think, herel2024thinking, pfau2024let, wang2024guiding} and seems to have emerged in frontier LLMs \citep{ryan_greenblatt_recent_2025, ryan_greenblatt_recent_2026, baherwani2026llmreasoningvisiblechainofthought, brauer2026reading}. Similarly, hidden reasoning on EoS tokens appears to emerge from the standard training procedure of much smaller masked diffusion models. This is possible because trailing sequences of EoS are in-distribution, as they are used as padding that is not masked out during training. Thus, they do not impact the output distribution negatively as opposed to other trailing token chains and are therefore effective across tasks. In addition, the bidirectional attention allows the diffusion models to attend to these trailing token positions.

Moreover, it has been shown that chains of thought are often not faithful to the inner workings of LLMs \citep{turpin2023language}. \citet{pfau2024let} highlight this tension by using dots as fully uninterpretable reasoning tokens. This problem is exacerbated in our experiments as diffusion LLMs not only seem to hide the true content of their reasoning, but also hide that they perform auxiliary reasoning altogether since the EoS tokens are decoded into empty strings.

\section*{Limitations}
In this work, we rely on a broad definition of reasoning that is used in the literature and subsumes the use of any intermediate representations, and in the case of chain-of-thought, intermediate tokens to affect model predictions \citep{wei2022chain}, as we've found repeatedly in our experiments, and do not aim to characterize reasoning in a deeper way. Furthermore, we do not assume that the reasoning steps of the models are human-like or easily human-interpretable. We also have not decoded what specific intermediate circuits or sub-steps are computed within the models' hidden states, since our goal was not to mechanistically understand how the reasoning behavior is realized in diffusion LLMs but to analyze why the seemingly innocuous hyperparameter, generation length, and tied to it, the generation of EoS tokens, have an impact on the model performance.

Further, our evaluation focuses exclusively on three recent masked diffusion language models: LLaDA1.5, LLaDA-2.0-mini, and Dream-v0. While our results consistently suggest that fully bidirectional attention is more favorable to the emergence of latent reasoning in EoS tokens than block-causal attention, the limited selection of models does not allow for a more in-depth comparison of architectural choices. This would require a larger selection of masked diffusion LLMs with different setups regarding their architecture and training procedure. However, at least for now, there are only a few publicly available large diffusion LLMs, which prevented a more systematic comparison of different model architectures.
\section*{Ethical Considerations}
This research focuses on the interpretability of diffusion language models. As the study is not investigating any downstream uses of language models, we do not foresee any immediate ethical concerns or negative societal consequences arising from this work.

\paragraph{LLM use.} We did not use LLMs in preparing this paper beyond fixing typos or grammar in individual paragraphs.

\section*{Reproducibility Statement}
The experiments can be conducted using one H100 GPU. We provide full details of the model hyperparameters and prompts in the appendix. The datasets, implementation, and evaluation scripts are  available on GitHub: \url{https://github.com/minaolensarah/thinking_eos-by-eos}

\section*{Acknowledgments}
This research has been funded by the Vienna Science and Technology Fund (WWTF) [10.47379/VRG23007] ``Understanding Language in Context.''

\bibliography{custom}

\appendix

\section{Other potential reasoning tokens}\label{otherTokens}
This paper focuses on masked diffusion LLMs' ability to leverage EoS tokens for hidden reasoning, grounded in the observation that the models produce such token chains when the generation length is set sufficiently high. However, just because models generally produce these specific tokens may not imply that latent reasoning is constrained to the representations of EoS tokens. As a control experiment, we therefore inspect whether diffusion models make use of other token repetitions at the end of the generation window for reasoning. We repeat the controlled prompting experiment (Exp.~2, Section~\ref{exp2}) and the causal intervention (Exp.~3, Section~\ref {exp3}) with dots (`.'), whitespaces(` '), and randomly sampled tokens (using two different random seeds), replacing the EoS tokens at trailing positions. Since LLaDA2.0-mini exhibited very inconsistent behavior in the previous experiments, we limit this control experiment to Dream and LLaDA1.5. 

Note that this control experiment is not compatible with the unconstrained generation setup (Exp.~1, Section~\ref{exp1}) as models always use EoS tokens to pad outputs. Any other tokens that may be used for latent reasoning, therefore, have to be explicitly inserted into the start state, as is the case for the setup of the controlled prompting and the intervention experiment.

\subsection{Controlled Prompting (Exp.~2)}\label{otherTokens_exp2}

\begin{figure*}[h]
    \centering
    \includegraphics[width=0.9\linewidth]{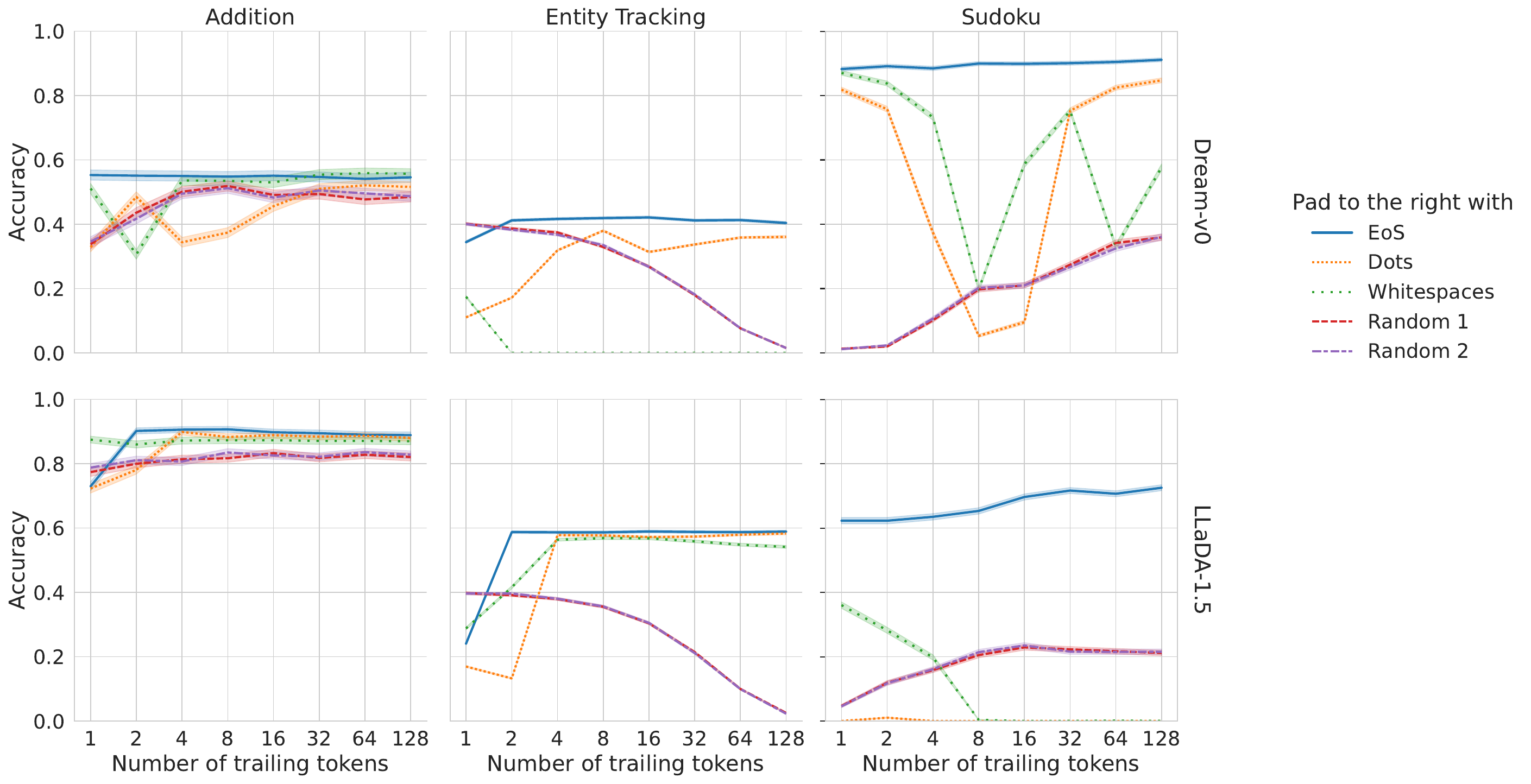}
    \caption{Accuracies with standard error of appending dots, whitespaces, and random token chains as potential reasoning tokens in the setup of Experiment~2.}
    \label{fig:otherTokens_exp2}
\end{figure*}
We add an increasing number of potential reasoning tokens at the end of the initial state of the stepwise generation, repeating the setup of Experiment~2. Figure \ref{fig:otherTokens_exp2} shows the accuracies across models, tasks, and token identities. Overall, EoS tokens achieve the highest and most stable performance compared to the other tokens. Dots and whitespaces match EoS performance in some cases, such as Dream on \textsc{Entity Tracking}, but lead to extreme drops in others, such as \textsc{Sudoku} for both models. Random token chains lead to lower accuracies, except for \textsc{Entity Tracking} for exactly one added token, and show heterogeneous trends. Across models, on \textsc{Addition} and \textsc{Sudoku}, an increase in padded random tokens leads to an increase in performance, while on \textsc{Entity Tracking} the accuracy declines with the number of padded tokens. Similarly, \citet{baherwani2026llmreasoningvisiblechainofthought} find that whether an autoregressive frontier model can exploit a specific filler token depends on the model-task pairing.

A qualitative analysis shows that, especially on the Sudoku task, the other tokens lead to textual outputs such as "Unfortunately, this Sudoku puzzle is unsolvable." This highlights the contending functions of the additional tokens as reasoning space but also as carriers of semantic information. Any token chain increases the reasoning capacity, which can lead to improvements in a task, but this only reliably works if the token is expected given the training data distribution. Otherwise, the benefit of added computational space seems to be overruled by the additional lexical information of the appended tokens, which influences which other tokens the model will generate. For bidirectional diffusion LLMs, EoS tokens appear to lie at a sweet spot where they can be utilized as computational space without the identity of the token affecting the other tokens to generate.

\subsection{Causal Intervention (Exp.~3)}\label{otherTokens_exp3}
\begin{figure*}[h!]
     \begin{subfigure}[b]{\textwidth}
        \centering
        \includegraphics[width=\linewidth]{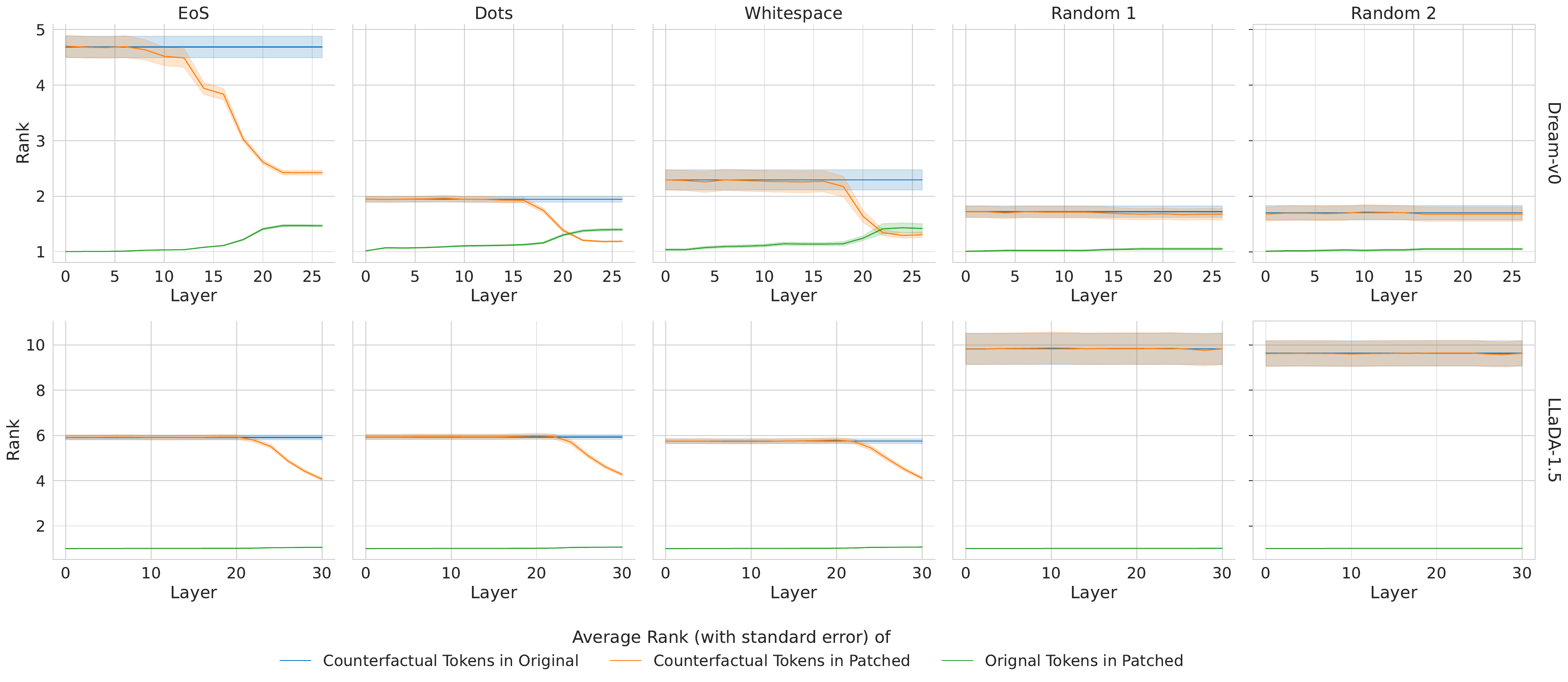}
        \caption{Addition.}
    \end{subfigure}
    \hfill
     \begin{subfigure}[b]{\textwidth}
        \centering
        \includegraphics[width=\linewidth]{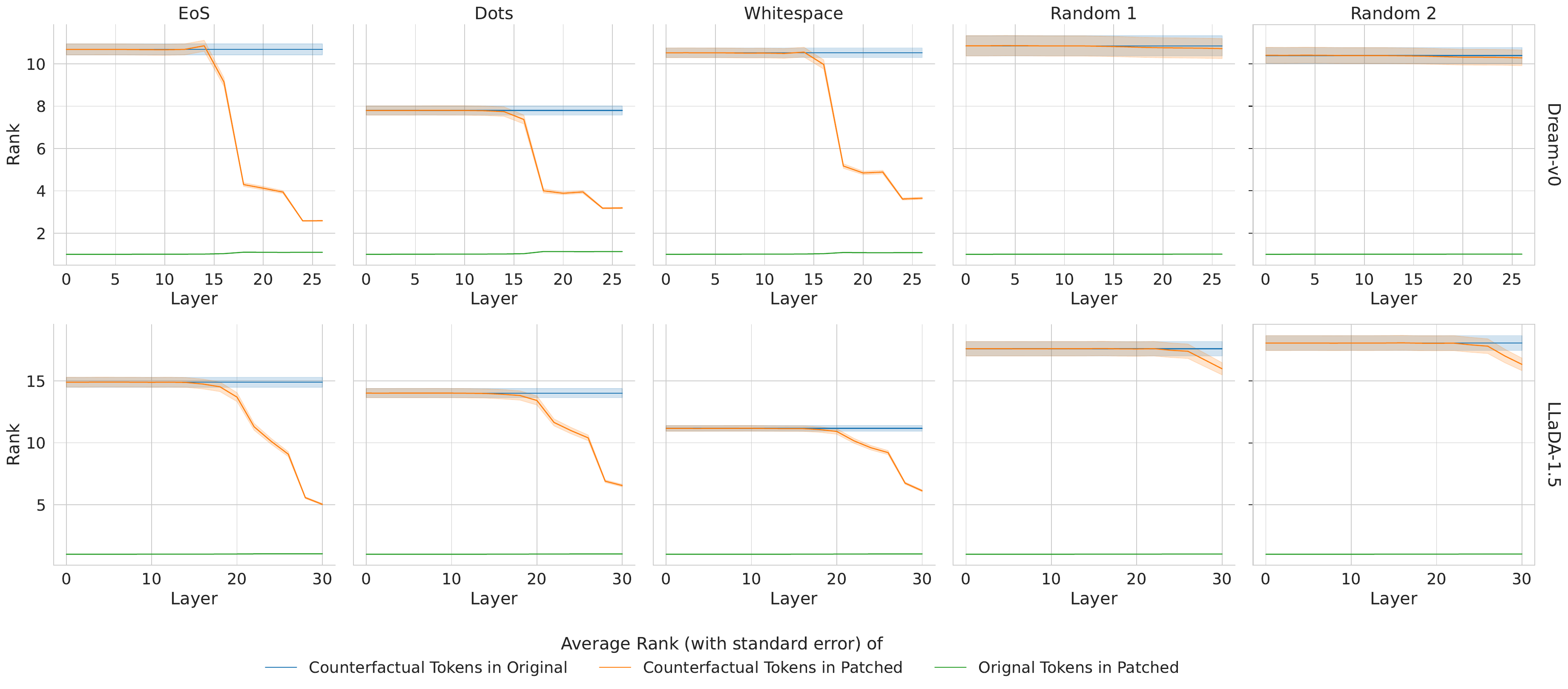}
        \caption{Entity Tracking.}
    \end{subfigure}
    \hfill
     \begin{subfigure}[b]{\textwidth}
        \centering
        \includegraphics[width=\linewidth]{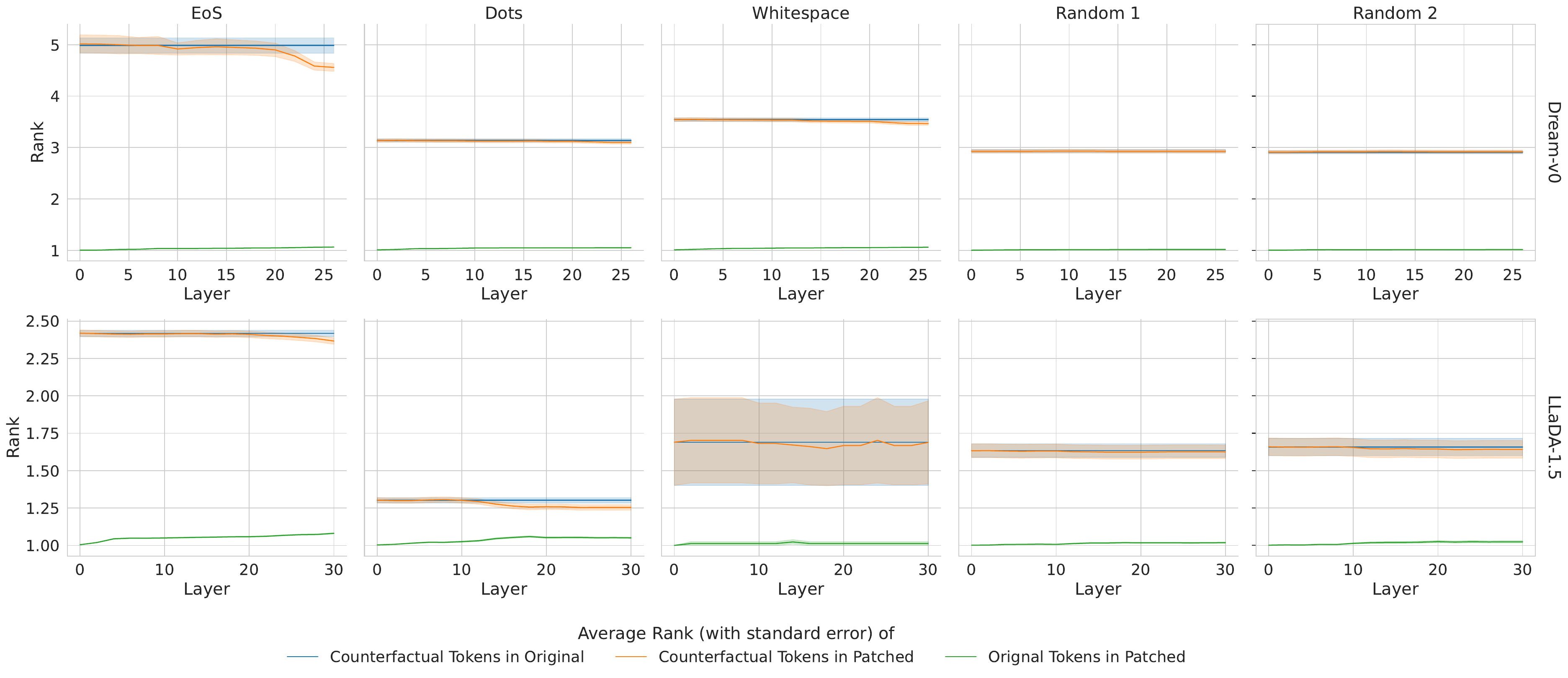}
        \caption{Sudoku.}
    \end{subfigure}
    \caption{Results of the intervention experiment (Exp.~3) executed with dots, whitespaces, and random token chains as potential reasoning tokens.}
    \label{fig:patching_otherTokens}
\end{figure*}

We repeat the activation patching experiment in which we transfer the hidden states of the 64 trailing EoS tokens with dots, whitespaces, and random tokens with two random seeds at the patched positions. Figure \ref{fig:patching_otherTokens} displays the respective token ranks in the output distributions per task, model, and filler identity. The difference between the rank of the counterfactual tokens in the original generation (blue) and their rank after patching up until a specific layer (orange) indicates whether the patched representations affect the output distribution. A decrease in the rank suggests that the hidden states contain additional latent computations that are being used by the models.

Patching EoS tokens impacts the output distribution across tasks and models except for LLaDA1.5 on \textsc{Sudoku}. On the contrary, random token chains as potential reasoning tokens lead to hardly any changes to the rank of the counterfactual tokens and a large amount of excluded invalid generations, except for LLaDA1.5 on \textsc{Entity Tracking}, where the rank of the counterfactual tokens decreases by a small margin. In between these two extremes lie dots and whitespaces as potential reasoning tokens. Patching their hidden states leads to changes in the output distribution on the \textsc{Entity Tracking} and \textsc{Addition} tasks for both models, indicating that the models use the filler tokens for additional compute. In the case of patching Dream on the \textsc{Addition} data, the original and counterfactual token ranks change their ordering; thus, the effect of the patching is visible in the final generation. However, the tokens change between ranks 1 and 2, indicating that the original and counterfactual generations were very similar before patching. A qualitative analysis shows that the model produces incomplete generations, such as one-digit answers padded with dots, for addition problems with longer solutions. These padding tokens result in overall smaller ranks for the generations compared to correct outputs.   

To sum up, we find that representations of filler tokens other than EoS appear to be also used for latent computations, but their effectiveness is very dependent on the combination of model and task. In contrast, interventions on EoS tokens lead to changes in the output distribution more consistently.

\section{Exp.~1: Measured output lengths} \label{appendix:exp1}
\begin{figure*}[p]
    \centering
    \includegraphics[width=0.9\linewidth]{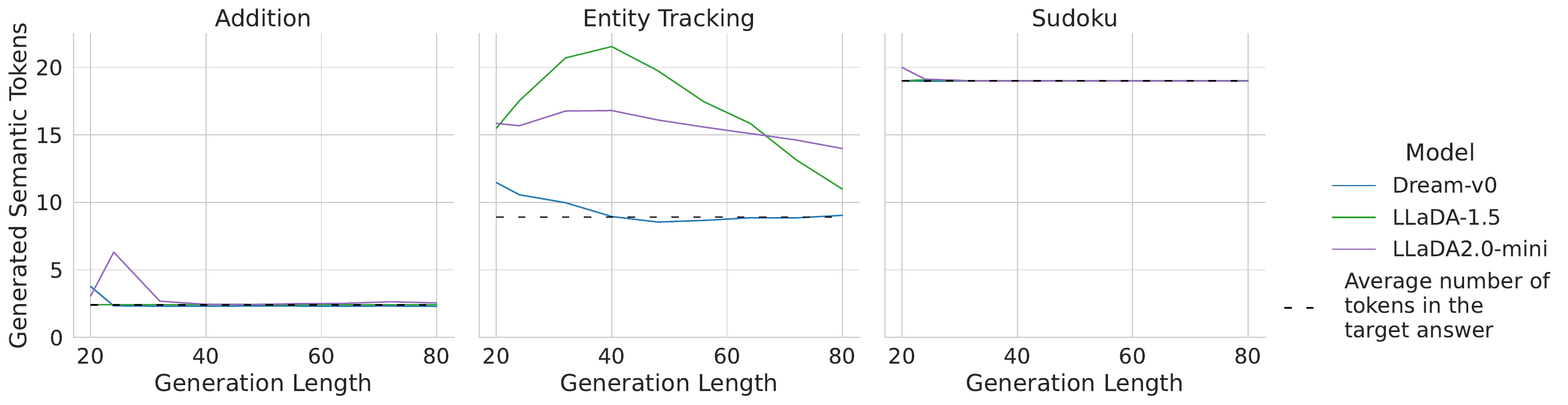}    
    \caption{The number of semantic tokens generated over the predefined generation length. The remaining positions are padded with EoS tokens by the model. The dashed lines indicate the average length of the target answer.}
    \label{fig:observations:length}
\end{figure*}
Figure~\ref{fig:observations:length} shows that for \textsc{Addition} and \textsc{Sudoku}, the number of semantic tokens stays constant across generation lengths. Consequently, the number of EoS tokens increases to fill the remaining positions. For \textsc{Entity Tracking}, the number of semantic tokens decreases when the generation length increases. This is due to a common error pattern across models. They tend to predict too many objects as contents of the target box, which results in them generating too many semantic tokens. For this task, we further find that the higher the generation length, the better the model performance, because the models predict the correct number, i.e., fewer, objects, leading to a decline in the number of semantic tokens.

\section{Exp.~2: Accuracies over problem difficulty} \label{app:exp2}
\begin{figure*}[p]
    \centering
    \includegraphics[width=0.9\linewidth]{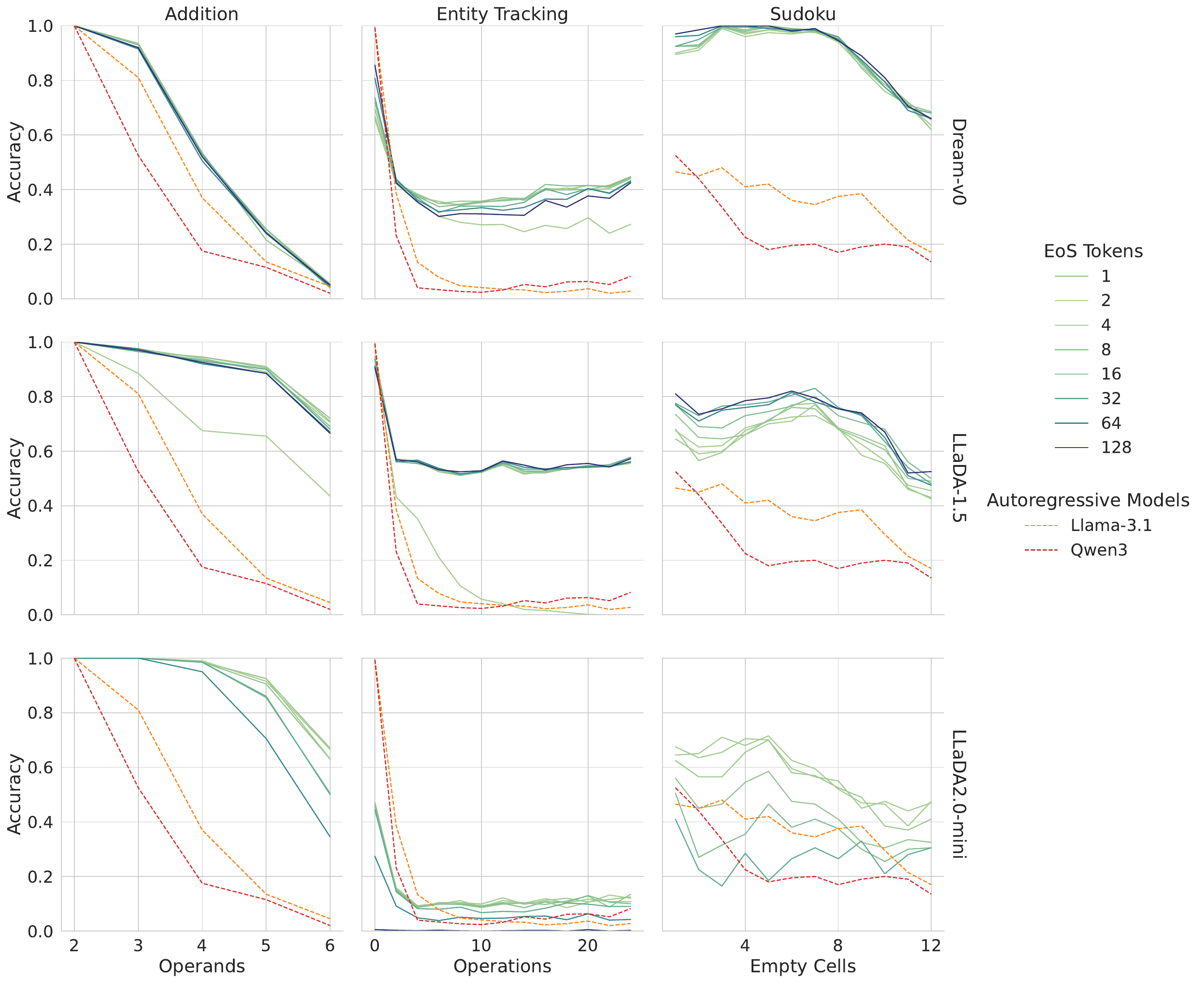}
    \caption{Detailed results of the controlled prompting experiment (Exp.~2). The accuracy of the diffusion models across task-specific difficulty levels and trailing EoS tokens. We show the performance of the autoregressive LLMs as dashed lines.}
    \label{fig:eos:acc_over_diff}
\end{figure*}
As for the first experiment, we report the accuracies across task-specific levels of difficulty in Figure \ref{fig:eos:acc_over_diff}, such as the number of operations on the target box to follow in the entity tracking task. For example, LLaDA1.5 on \textsc{Sudoku} has a more stable performance for up to 10 empty cells when more EoS tokens are appended to the start state. In general, the effect is weaker than when increasing the generation length.
\begin{table*}[p]
    \centering
    \begin{tabular}{l|l|l|l}
    \toprule
        \textbf{Dataset} & \textbf{Counterfactual} & \textbf{Size} & \textbf{Detailed Size (200 each)} \\
        \midrule
        Addition & swap operators & 1000 & 2-6 operands\\
        Entity Tracking   & different box & 12900 & 2, 4...30 global operations x 0, 2, 4...  \\
        \citep{kim-schuster-2023-entity} & & & max(global ops, 24) local operations\\
        Sudoku \citep{Sudoku4LLM} & other Sudoku & 2400 & 1-12 empty cells \\        
        \bottomrule
    \end{tabular}
    \caption{Dataset statistics.}
    \label{tab:datasetstats}
\end{table*}
\section{Hyperparameters}
\label{app:hyperparameters}
\paragraph{Exp.~1: Prompting Experiment} We use zero-shot prompts for \textsc{Addition} and \textsc{Entity Tracking}, and a two-shot prompt for \textsc{Sudoku} to introduce the models to our format of the puzzles. In all setups, we instruct the models not to generate any verbose reasoning steps but to immediately produce the answer. The block length and the number of decoding steps are equal to the generation length, which we increase from 20 to 80 tokens. The autoregressive baselines are run with their default hyperparameters and generation length 20.
\paragraph{Exp.~2: Controlled Prompting Experiment} In the start state, we set the number of masks to the maximum length of the target answer of the dataset, 4 for \textsc{Addition}, 22 for \textsc{Entity Tracking}, and 19 for the \textsc{Sudoku} formatted with line breaks. The masks are padded with 1, 2, 4, 8...128 EoS tokens. On the \textsc{Sudoku} task, if applicable to the model, we insert an end-of-turn-token instead of the first padded EoS token. The number of decoding steps is equal to the number of masks; hence it remains constant across runs and numbers of padded EoS tokens.
\paragraph{Exp.~3: Intervention Experiment} For a target layer $l$, we patch the hidden representations of the positions of the EoS tokens after each layer $l_i\leq l$ of the counterfactual run into the original run after layer $l_i$. We do not run stepwise decoding but predict all masks at once. We use 64 trailing EoS tokens. For \textsc{Addition} we use four masks, for \textsc{Entity Tracking} we add 22 masks, and for \textsc{Sudoku} we insert the Sudoku in the assistant's answer with masks at the positions of the empty cells.
\paragraph{Exp.~4: Experiment on more naturalistic benchmarks} As for Experiment~2, we set the number of decoding steps equal to the number of masks, but we use temperature 0.1 for all diffusion models following Dream's evaluation protocol. On \textsc{GSM8K} we use a 5-shot prompt with reasoning traces but instruct the model to only give the answer, and the generation starts from 10 masks. For \textsc{two-hop} we use six masked positions. The autoregressive baselines are run with their default hyperparameters and generation length equal to the number of masks.

\section{Dataset statistics} \label{datastats}

\paragraph{Addition, Entity Tracking, and Sudoku} See Table~\ref{tab:datasetstats}. As entity tracking data, we generate scenarios with 7 boxes containing a maximum of 3 objects and an expected number of 2 objects. The problem descriptions contain up to 30 operations, of which zero to 24 concern the box that the model is queried for. 
\paragraph{GSM8K \citep{cobbe2021gsm8k}.} We use the full test set comprising 1319 examples. 
\paragraph{Two-hop \citep{biran-etal-2024-hopping}.} We exclude descriptions that ask for a current head of state, as this knowledge depends on the models' cut-off dates. Moreover, we only use data points with a tokenized answer length shorter than 6. Then, we test the models on the one-hop parts of the description and filter model-specific test sets that only include two-hop questions whose parts the models answered correctly in isolation. The test sets consist of 4058 examples for Dream, 2776 for LLaDA1.5, 5107 for LLaDA2.0-mini, 5112 for LLama3.1, and 5495 for Qwen3.

\section{Prompts} \label{sec:appendix:prompts}
\paragraph{Addition}
\begin{Verbatim}[commandchars=\\\{\}, breaklines=true]
\textbf{System:}
Answer the question only with the number that is the final result. Do not give any additional explanation.
\textbf{User:}
What is the result of 99+18-23?
\end{Verbatim}

\paragraph{Entity Tracking}
\begin{Verbatim}[commandchars=\\\{\}, breaklines=true]
\textbf{System:}
Answer the question but do not give any additional explanation. 
\textbf{User:}
Box 0 contains the key and the plant, Box 1 contains the dish, Box 2 contains the block and the shell, Box 3 contains the brick and the flower and the string, Box 4 contains nothing, Box 5 contains the card, Box 6 contains the cash and the guitar and the wire. Remove the brick from Box 3. Put the mirror into Box 3.
What does Box 0 contain?
\end{Verbatim}

\paragraph{Sudoku} The system prompt contains the explanation of the Sudoku rules by \citet{Sudoku4LLM}.
\begin{Verbatim}[commandchars=\\\{\}, breaklines=true]
\textbf{System:}
4x4 Sudoku Rules:
    - The grid is 4x4 in size.
    - Each row, column, and 2x2 sub-grid must contain the numbers 1 to 4 exactly once.
    - Some cells are pre-filled, and the player must fill in the rest.
Only provide the solved sudoku grid as a string of digits. Do not provide any additional explanation or text.

\textbf{User:}
Solve the following Sudoku puzzle:
0000
0040
4312
0200

\textbf{Assistant:}
3421
2143
4312
1234

\textbf{User:}
Solve the following Sudoku puzzle:
0400
3014
2300
4032

\textbf{Assistant:}

1423
3214
2341
4132

\textbf{User:}
Solve the following Sudoku puzzle:
4312
0134
1423
3241
\end{Verbatim}
\paragraph{GSM8K} The 5-shot demonstrations are taken from \citet{eval-harness}.
\begin{Verbatim}[commandchars=\\\{\}, breaklines=true]
\textbf{System:}
Given the following problem, give a final answer to the problem. Give only the result as an answer. Do not write any further explanations. Your response should be ""The final answer is [answer]"" where [answer] is the response to the problem.
\textbf{User:}
Here are some examples of problems and how to arrive at the final answer. In your answer only give the answer and not the reasoning.
Problem: There are 15 trees in the grove. Grove workers will plant trees in the grove today. After they are done, there will be 21 trees. How many trees did the grove workers plant today?
Reasoning: There are 15 trees originally. Then there were 21 trees after some more were planted. So there must have been 21 - 15 = 6. 
Answer: The final answer is 6
Problem: If there are 3 cars in the parking lot and 2 more cars arrive, how many cars are in the parking lot?
Reasoning: There are originally 3 cars. 2 more cars arrive. 3 + 2 = 5. 
Answer: The final answer is 5
Problem: Leah had 32 chocolates and her sister had 42. If they ate 35, how many pieces do they have left in total?
Reasoning: Originally, Leah had 32 chocolates. Her sister had 42. So in total they had 32 + 42 = 74. After eating 35, they had 74 - 35 = 39. 
Answer: The final answer is 39
Problem: Jason had 20 lollipops. He gave Denny some lollipops. Now Jason has 12 lollipops. How many lollipops did Jason give to Denny?
Reasoning: Jason started with 20 lollipops. Then he had 12 after giving some to Denny. So he gave Denny 20 - 12 = 8. 
Answer: The final answer is 8
Problem: Shawn has five toys. For Christmas, he got two toys each from his mom and dad. How many toys does he have now?
Reasoning: Shawn started with 5 toys. If he got 2 toys each from his mom and dad, then that is 4 more toys. 5 + 4 = 9. 
Answer: The final answer is 9

Problem: Janet’s ducks lay 16 eggs per day. She eats three for breakfast every morning and bakes muffins for her friends every day with four. She sells the remainder at the farmers' market daily for \$2 per fresh duck egg. How much in dollars does she make every day at the farmers' market?
Remeber, do not write down your reasoning, only give the answer.
\end{Verbatim}
\paragraph{Two-hop}
\begin{Verbatim}[commandchars=\\\{\}, breaklines=true]
\textbf{System:}
Provide only the name of the entity described by the statement. Do not give any explanation or reasoning trace.
\textbf{User:}
The capital of the country of citizenship of stephen harper is
\end{Verbatim}
\end{document}